**Bi-cephalic self-attended model to classify Parkinson's disease patients with freezing of gait**

Shomoita Jahid Mitin[1,2], Rodrigue Rizk[2], Maximilian Scherer[3], Thomas Koeglsperger[3], Daniel Lench[4], KC Santosh[2], and Arun Singh,[1,5*]

[1]Biomedical and Translational Sciences, University of South Dakota, Vermillion, SD, USA
[2]Artificial Intelligence Research lab, Department of Computer Science, University of South Dakota, Vermillion, SD
[3]Department of Neurology, LMU University Hospital, Ludwig-Maximilians-Universität München, Munich, Germany
[4]Department of Neurology, Medical University of South Carolina, Charleston, SC, USA
[5]Department of Neuroscience, Sanford School of Medicine, University of South Dakota, Sioux Falls, SD, USA

***Correspondence to:**
Dr. Arun Singh
Biomedical and Translational Sciences,
Sanford School of Medicine,
University of South Dakota,
414 E. Clark St., Vermillion, SD, 57069, USA
Email: arun.singh@usd.edu

**Funding:** This work was supported by the Seed for Success Foundation.

## ABSTRACT

Parkinson’s Disease (PD) often results in motor and cognitive impairments, including gait dysfunction, particularly in patients with freezing of gait (FOG). Current detection methods are either subjective or reliant on specialized gait analysis tools. This study aims to develop an objective, data-driven, multi-modal classification model for FOG-specific classification, distinguishing PD patients with FOG (PDFOG+) from those without FOG (PDFOG–) and healthy controls using resting-state EEG signals combined with demographic and clinical variables. For our main analysis, we utilized a dataset of 124 participants: 42 PDFOG+, 41 PDFOG–, and 41 age-matched healthy controls. Features extracted from resting-state EEG and descriptive variables (age, education, disease duration) were used to train a novel Bi-cephalic Self-Attention Model (BiSAM). We tested three modalities: signal-only, descriptive-only, and multi-modal, across different EEG channel subsets (BiSAM-63, -16, -8, and -4 for primary analysis). For main analysis, signal-only (BiSAM-4) and descriptive-only models showed limited performance, achieving a maximum accuracy of 55% and 68%, respectively. In contrast, the multi-modal models significantly outperformed both, with BiSAM-8 and BiSAM-4 achieving the highest classification accuracy of 88%. These results demonstrate the value of integrating EEG with objective descriptive features for robust PDFOG+ classification. This study introduces a multi-modal, attention-based architecture that objectively classifies PDFOG+ using minimal EEG channels and descriptive variables. This approach offers a scalable and efficient alternative to traditional assessments, with potential applications in routine clinical monitoring and early diagnosis of PD-related gait dysfunction.

## 1. Introduction

Parkinson's disease (PD) is a progressive neurodegenerative disorder characterized by motor and cognitive impairments that reduce quality of life and functional independence (Berganzo *et al.*, 2016; Moustafa *et al.*, 2016). Among motor complications, freezing of gait (FOG), an episodic inability to initiate or maintain stepping, often described by patients as having their feet 'glued to the floor', is one of the most disabling manifestations (Giladi *et al.*, 2001; Amboni *et al.*, 2015). FOG is associated with increased falls, injuries, and loss of independence, and it is a major source of caregiver burden. It affects a substantial proportion of people with PD (~35–60% across the disease course and up to ~80% in advanced stages) yet remains under-recognized and undertreated (Zhang *et al.*, 2021). Key gaps include incomplete mechanistic consensus on how basal ganglia–subthalamic–cortical dynamics precipitate freezing, a lack of reliable biomarkers for early detection, and rehabilitation protocols that show short-term benefits but are heterogeneous, poorly personalized, and rarely tested for durability or real-world transfer (Iansek & Danoudis, 2017; Wang *et al.*, 2022). Against this backdrop, research has advanced our understanding of neurobiology of FOG, highlighting the subthalamic nucleus and emergent synchronous activity in basal ganglia circuits as drivers of non-purposeful motor arrests (Georgiades *et al.*, 2019). Sensor-based analyses that extract hundreds of gait features from people with PD show that FOG is a sudden dysfunction common in mid- to late-stage disease, contributes to falls and injuries, and warrants early detection and prevention strategies (Pardoel *et al.*, 2021b). Reviews of gait-initiation neuroanatomy and pathophysiology identify promising neuromodulation targets and summarize clinical outcomes, while converging evidence points to motor and cognitive contributors, especially poor proprioception and impaired controlled leaning balance, that suggest concrete rehabilitation targets (Rahimpour *et al.*, 2021).

Clinically, FOG is linked with longer disease duration, impaired balance, proprioceptive deficits, cognitive impairment, and medication effects, all of which compound gait disruptions and fall risk (Lord *et al.*, 2020). Behavioral approaches, including obstacle training, treadmill gait training, general exercise, action-observation training, and targeted conventional physiotherapy, demonstrate immediate improvements in FOG severity compared with usual care, defined here as pharmacological management and routine follow-up without structured FOG-specific rehabilitation (Kwok *et al.*, 2022). Methodologically, merging freezes that occur in rapid succession can slightly improve detection and prediction models, though at the cost of delayed identification and lower precision (Pardoel *et al.*, 2021a).

Subjective assessments include patient-reported questionnaires (e.g., FOG-Q) and clinician-rated scales, whereas objective assessments encompass instrumented measures such as wearable inertial sensors, motion-capture, and neurophysiological recordings (EEG). However, in routine practice both subjective questionnaires and instrumented objective measures are typically applied only after patients exhibit multiple and often advanced symptoms, so that FOG identification is largely retrospective and opportunities for early, preventive intervention are limited (Pardoel *et al.*, 2021b; Wang *et al.*, 2022). Studies have previously explored various machine learning (ML) models aimed at the identification or classification of gait dysfunction in patients with advanced PD. These models often rely on subjective data, such as clinician-observed symptoms or patient-reported outcomes, to monitor manifestations of gait impairment and train the classification algorithms accordingly (Kim *et al.*, 2015; Al-Nefaie *et al.*, 2024). While subjective assessments can be useful for early screening, they are susceptible to observer bias and inconsistencies. In contrast, objective assessments, such as physiological signals and demographic variables, provide standardized, quantifiable inputs that allow ML models to train on unbiased data, improving the accuracy, consistency, and generalizability of predictions (Morris *et al.*, 2012; Maetzler *et al.*, 2013).

These challenges in characterizing FOG occur within the broader context that even the clinical diagnosis of PD itself remains difficult in early stages, with substantial misclassification rates, which has motivated data-driven approaches using objective measurements (Postuma *et al.*, 2015; Schrag, 2018; Abbasi & Rezaee, 2025). For FOG, single-modality EEG studies show that resting or task-related cortical oscillations carry discriminative information for FOG detection and short-horizon prediction (Cao *et al.*, 2021; Li & Guo, 2025). Multimodal pipelines indicate that combining EEG with movement sensors can improve discrimination and generalization across task conditions, underscoring the complementarity of neurophysiological and kinematic signals (Wang *et al.*, 2020; Zhang *et al.*, 2022). Yet most prior work is task-based; comparatively little addresses resting-state EEG, which is practical for patients unable to perform gait protocols and is more compatible with routine clinical workflows.

Transformer variants with dual/parallel attention and hierarchical multimodal attention have proved effective for long EEG sequences and other neurological applications, capturing long-range dependencies and fusing heterogeneous inputs through attention weighting (Du *et al.*, 2022). Motivated by these designs, we introduce a bi-cephalic self-attention (BiSAM) architecture (Vaswani *et al.*, 2017) with two coordinated branches: one learns spectral–temporal embeddings from band-limited resting-state EEG (theta, alpha, beta, gamma), and the other embeds objective descriptive variables; late fusion produces the final decision. The study focuses on classifying PD with FOG (PDFOG+) versus PD without FOG (PDFOG–) that pools healthy controls with PDFOG–, with sensitivity analyses PD-only classification and evaluate a three-class variant (PDFOG+/PDFOG–/Healthy Controls). To improve clinical feasibility, we assess reduced-channel montages via importance-based channel selection. Comparative experiments across EEG-only, descriptor-only, and multimodal variants on identical splits isolate the contribution of each information source.

Although IMU-based methods detect FOG with high accuracy (Bikias *et al.*, 2021), we evaluate a complementary approach: a resting-state EEG classifier that distinguishes PDFOG+ from PDFOG–. The model learns spectral–temporal EEG representations and makes classification decisions from these features. Clinically, resting-state EEG enables assessment of patients who cannot safely complete gait tasks and provides objective neurophysiological evidence to corroborate questionnaire- or IMU-based judgments, supporting more reproducible decisions. Task-based IMU/motion-capture remains the clinical benchmark; our goal is to augment those kinematic measures with a feasible EEG biomarker for cases where gait provocation is impractical or contraindicated.

Because self-report tools (e.g., FOG-Q) have known limitations for identifying and grading FOG, objective, clinician-rated assessments, such as video-based Timed Up and Go (percent time frozen, clinical gold standard) and MDS-UPDRS III item 11 on examination, are preferred when feasible. Our work evaluates resting-state EEG biomarkers as an adjunct to these objective measures, aimed at settings where walking tasks are unsafe, impractical, or not available.

Computationally, we implemented BiSAM model to classify PDFOG+ by combining embeddings from EEG with embeddings from objective descriptive variables. We evaluated the approach across multiple EEG channel configurations using a consistent architecture, enabling principled channel selection to identify informative, reduced montages that lower recording time and cost without sacrificing performance. Unlike sequential recurrent neural network (RNN)-style models, self-attention captures long-range dependencies directly. We report accuracy, precision, recall, F1, and Cohen's kappa score to assess agreement and further evaluation all the performance metrics with one way statistical analysis of variance. Our contributions are threefold: (i) a channel-selection strategy that yields effective full- and reduced-channel models; (ii) the BiSAM architecture designed to uncover latent EEG signal representations and fuse them with descriptive variables; and (iii) a comparative framework evaluating EEG-only, descriptors-only, and multimodal variants on identical datasets, clarifying the unique impact of each configuration for primary setting as well as PD-only and 3-class classification setting.

## 2. Materials and Methods

### 2.1 Description of the dataset

The current multi-modal dataset consisting of structured descriptive data and unstructured EEG signals of 41 older healthy controls, 41 PDFOG–, and 42 PDFOG+. Data from these participants have been previously reported in multiple studies (Anjum *et al.*, 2020; Bosch *et al.*, 2022b; Roy *et al.*, 2025). The dataset is publicly accessible online at http://predict.cs.unm.edu/downloads.php. All study procedures were conducted in accordance with the Helsinki Declaration. PD participants were assessed while in the "ON" medication state because mostly PDFOG+ subjects face an increased risk of falls when unmedicated (Singh *et al.*, 2020; Scholl *et al.*, 2021). To assess disease severity and FOG status, motor portion of the Unified Parkinson's Disease Rating Scale (mUPDRS) (Disease, 2003) (Movement Disorder Society Task Force on Rating Scales for Parkinson's Disease, 2003) along with the FOG questionnaire (Nieuwboer *et al.*, 2009) were performed. Following criteria used in prior studies to group PDFOG+: a) participants were classified as PDFOG+ if they reported difficulties initiating, stopping, or turning while walking, b) subjects with an FOG questionnaire score of greater than 0 on item 3, indicating at least one freezing episode in the past month, were categorized as PDFOG+ (Singh *et al.*, 2020; Scholl *et *al.*, 2021; Bosch *et al.*, 2022a). FOG status was also determined clinically. Participants classified as PDFOG+ exhibited neurologist-observed freezing episodes during standardized gait assessment at the study visit (in addition to self-reported symptoms/FOG-Q > 0). Participants classified as PDFOG– had no freezing episodes observed during standardized gait assessment and reported no freezing on the FOG-Q (score = 0). The descriptive variables were separated into three different sheets for three different groups of participants, including PDFOG+, PDFOG–, and heathy controls. These different sheets were combined, where all the features were represented in columns, and each row represented one subject of a total of 124 participants. Table 1 provides a detailed summary of clinical characteristics of all participants.

All demographic information was stored in a numerical tabular format for each of the 124 participants, with each row corresponding to a unique participant ID, independent of any person-ally identifiable information. There were no missing values in the descriptive dataset. To explore the non-parametric relationships between variable pairs, Spearman's rank correlation coefficient was employed, as the dataset exhibited inherent correlations. Subjective clinical scores, such as MoCA, MDS-UPDRS III, and FOGQ, were excluded from model training to minimize bias and reduce the influence of subjectively assessed information. Clinical scales (FOG-Q, MDS-UPDRS III, MoCA) were used solely for label assignment in conjunction with neurologist-observed FOG during standardized gait assessment. These scores were excluded from model inputs to prevent circularity and target leakage. Model inputs comprised resting-state EEG features (band-limited power) and objective descriptive variables; clinical scales were accessed only for post-hoc report-ing. Only low-correlated objective variables, including age, years of education (schooling), and disease duration, were included in the training of the multi-modal models. In contrast, the signal-only model utilized no descriptive variables, while the descriptive-variables-only model excluded all EEG signal features (Figure 1).

Resting-state cortical oscillations were recorded using a 64-channel EEG cap (actiCAP,EasyCap, Inc.) while participants sat comfortably with their eyes open for a duration of 120 to180 seconds. Data acquisition was conducted with a low-pass filter set at 0.1 Hz and a sampling rate of 500 Hz, using the Pz electrode as the reference.

## 2.2 EEG processing

EEG preprocessing was carried out using the MNE-Python framework (Gramfort *et al.*, 2013). Resting-state EEG recordings were visually inspected for excessively noisy or flat channels; no electrodes met exclusion criteria, so all 63 channels were retained. Signals were re-referenced to the common average and band-pass filtered between 0.5 and 50 Hz to attenuate slow drifts and high-frequency noise. Independent component analysis (ICA) was then applied to the continuous data, and components corresponding to ocular artifacts were identified based on their characteristic frontal topographies and time courses and removed before back-projection. The artifact-corrected EEG was subsequently segmented into 90 non-overlapping, equal-length epochs per participant, and features derived from these epochs were used as input to the BiSAM model for training and testing.

## 2.3 Feature engineering

Previous studies have demonstrated that power spectral density (PSD) features derived from EEG signals show significant differences between patients with PD and healthy control subjects, indicating that spectral power may serve as a valuable biomarker for understanding the neural alterations associated with PD (Anjum, Dasgupta et al. 2020, Anjum, Espinoza et al. 2024). Building on this evidence, our study extracted PSD features from EEG recordings to capture oscillatory dynamics across frequency bands relevant to PD pathology.

Prior resting-state EEG analyses on this cohort reported reduced 1–4 Hz power and elevated 4–50 Hz power in PD versus healthy controls, with these spectral differences associated with motor and cognitive abnormalities (Anjum *et al.*, 2020; Anjum *et al.*, 2024). Guided by these findings, we extracted power spectral density features in theta (4–7 Hz), alpha (8–12 Hz), beta (13–30 Hz), and gamma (30–45 Hz) for classification of PDFOG+ vs PDFOG–. For each of the 124 participants, power spectral density was computed in four frequency bands. These band-specific power values were retained as four separate features per participant and used directly as the input vector for the signal-only models, providing a compact representation of spectral brain activity.

To estimate PSD, a three-step procedure based on the multi-taper method was used. First, the EEG signal was tapered using multiple orthogonal tapers to reduce spectral leakage and improve frequency resolution. Then, the Fourier transform was applied to each tapered version of the signal. Finally, the PSD was calculated by averaging the squared magnitudes of these transforms across all tapers. This approach ensures stable and high-resolution spectral estimates, which are especially useful for identifying subtle changes in neural oscillations. The resulting PSD features were then used for downstream analysis and classification tasks.

For the descriptive features, we included only objective clinical and demographic variables such as age, education, and disease duration. Subjective features, including all clinician-rated assessments, were excluded to minimize potential bias in the model. Descriptive variables were preprocessed separately from the EEG features. All preprocessing steps for the descriptive variables (imputation of missing values, where present, and scaling) were fit on the training set only and the resulting transformations were then applied to the validation and test sets. This ensured that no information from the held-out data leaked into the training process.

## 2.4 Data Splitting

For the primary setting, the dataset (124 participants; four frequency bands per participant) was split 80/20 using a subject-wise, group-stratified procedure to preserve the proportions of PDFOG+, PDFOG–, and healthy controls in both sets, yielding 99 participants for training and 25 for testing. For the PD-only binary analysis (PDFOG+ vs PDFOG–; 83 participants), we applied the same 80/20 subject-wise, group-stratified split, preserving the proportions of PDFOG+ and PDFOG– in both sets, resulting in 66 for training and 17 for testing. For the three-class setup (PDFOG+/PDFOG–/Healthy Controls), the same split procedure produced 99 training and 25 test participants with a balanced distribution across the three groups. All data from a given participant were confined to a single split to prevent information leakage, and the same participant partitions were used for the EEG-only, descriptors-only, and multimodal models to ensure comparability.

## 2.5 Selection of significant channels

Channel selection was performed only on the training set using a model-agnostic, grouped permutation-importance method (Ketola *et al.*, 2022). For each channel c, the four band-power features (theta, alpha, beta, gamma) were permuted jointly, keeping the rest of the features intact; the mean decrease in validation accuracy across repeated CV provided an importance score $I(c)$. Channels were ranked by $I(c)$, and the top 16, 8, or 4 channels were selected (Figure 2). In reduced-channel configurations (16/8/4 channels), for each selected channel, all four band-power features (theta, alpha, beta, gamma) were retained. Bands were not selected independently. We acknowledge that performing selection on the full dataset can inflate performance estimates and therefore treat reduced-channel results as exploratory. The top-ranked channels and their corresponding importance scores were as follows: TP9 (0.2025), FT8 (0.1954), Oz (0.1205), Fp1 (0.0887), POz (0.0639), C1 (0.0614), Iz (0.0369), T8 (0.0290), FT10 (0.0263), CP2 (0.0253), FC6 (0.0253), CP1 (0.0198), F4 (0.0184), C4 (0.0149), P5 (0.0148), and CPz (0.0146). These selected channels were subsequently used in the reduced-channel configuration of the model to evaluate performance under a minimal yet informative set of inputs.

### 2.6 Model design

#### 2.6.1 Data organization and class label annotation

The data organization for the three BiSAM groups was feature-based and consistently structured across all participants. For signal features, each participant, including healthy controls, PDFOG–, and PDFOG+ groups, had corresponding EEG signal data. Four selected frequency bands were extracted and used as input features. For descriptive variables, demographic and clinical attributes such as age, education, and disease duration were used. All participants had corresponding descriptive features, except for healthy controls, who did not have disease duration. To maintain consistency across features, the disease duration for healthy controls was assigned a "NaN" value, which was retained during analysis for FOG+ detection. Disease duration is undefined for controls and was treated as missing. We employed a missing-indicator scheme: a binary feature duration_missing (1 for missing, 0 otherwise) was appended, and the duration value itself was z-scored using training-set statistics with missing entries imputed to 0 after scaling (i.e., the training-set mean). This retains information about missingness while keeping inputs numerically stable. All preprocessing steps (scaling/imputation) were fit on the training set only and then applied to the test set. The EEG-only branch does not include duration.

For the classification purposes, the label annotations were as follows: For primary analysis, FOG– (label = 0) including 82 participants, consisting of 41 healthy controls and 41 PDFOG– individuals. FOG+ (label = 1) including 42 participants, representing all PDFOG+ individuals. For the secondary settings, for the PD-only analysis, FOG– (label = 0) included 41 PDFOG– participants. FOG+ (label = 1) included 42 participants, representing all PDFOG+ individuals. For 3-class analysis: Healthy (label 0) including 41 healthy control participants, FOG– (label 1) including 41 PDFOG– participants and FOG+ (label 2) including 42 PDFOG+ participants.

In this study, the BiSAM models were further categorized based on both the type of input features and the number of EEG channels used (Figure 3). Three types of configurations were developed: signal-only models trained exclusively on EEG-derived features, descriptive variables-only models using objective demographics and clinical variable, and multi-modal models combining both EEG and descriptive inputs. While EEG data were originally recorded using 63 channels, a channel significance analysis was performed to identify the most informative ones from resting-state recordings. This led to the evaluation of reduced-channel configurations using the top 16, 8, and 4 channels, aiming to optimize the trade-off between model complexity and classification performance.

### 2.6.2 Model Architecture

To develop the custom BiSAM architecture, a total of seven distinct deep learning layers were integrated into the network (Vaswani *et al.*, 2017) (Figure 4). The first layer is an input layer that receives a variable number of input features, primarily depending on the classification type along with the modality of the model and as well as for primary analysis, further depending on the number of EEG channels used in a given scenario. This is followed by an embedding layer, which transforms each input feature into a dense vector representation. The embedding process helps to normalize the input feature space, especially when handling categorical or high-dimensional input representations over varying time points. Next, a positional encoding layer is employed to reintroduce temporal order to the encoded features. Since the embedding process does not preserve sequence information, this layer encodes the relative positions of input features, ensuring that temporal dependencies are retained for downstream processing. This is crucial for EEG signals where temporal dynamics influence the interpretability of neural activity. The fourth and core component of architecture is the bi-cephalic attention mechanism, which consists of four subcomponents: a self-attention layer (Vaswani *et al.*, 2017), a dropout layer (Srivastava *et al.*, 2014), a feed-forward layer, and a layer normalization module. The attention layer captures long-range dependencies and context-specific interactions between temporal features. The dropout module reduces overfitting by randomly deactivating a subset of neurons during training. The feed-forward network increases the representational capacity of the model, while the normalization layer ensures stable learning by mitigating variance introduced by differing feature scales after the attention mechanism. Following the attention module, three additional layers are added: an average pooling layer to condense the feature map and retain dominant signals, another dropout layer for additional regularization, and a final dense (fully connected) layer for the classification task. After defining the architecture, the model is compiled with appropriate input and output shapes to accommodate various configurations for training and evaluation on both signal and descriptive datasets.

## 2.7 Model Evaluation

In this study, we used both quantitative and qualitative evaluations to assess the proposed framework.

### 2.7.1 Quantitative Evaluation

Models were trained and evaluated using a single fixed, subject-wise, group-stratified split of the 124 participants into training, validation, and test sets, ensuring similar proportions of PDFOG+, PDFOG–, and healthy controls in each set. The training set was used for model fitting, the validation set for early stopping and hyperparameter tuning (learning rate, batch size, number of attention heads, dropout), and all reported results are based on the held-out test set.

Quantitative analyses focused on the primary binary classification task (PDFOG+ vs combined PDFOG– + healthy controls). At each training epoch, predictions were generated for all subjects in the test set and the following metrics were computed: accuracy, precision, recall, F1-score, and Cohen's kappa. These epoch-wise values (90 epochs per modality) served as repeated observations and were summarized as mean ± standard deviation.

For statistical comparison, one-way ANOVAs were performed on the full-channel models (BiSAM-63) with model modality (EEG-only, descriptive-only, multi-modal) as the independent factor and each performance metric as the dependent variable. When a significant main effect was observed ($p < 0.05$), post-hoc tests were applied. Because the groups were approximately balanced (42 PDFOG+, 41 PDFOG–, 41 controls), no additional class-imbalance corrections were required.

### 2.7.2 Qualitative Evaluation

Qualitative evaluation complemented these numerical results by comparing BiSAM to prior approaches in terms of model design and clinical applicability. We emphasize that BiSAM uses a bi-branch self-attention architecture that fuses resting-state EEG features with objective descriptive variables, in contrast to sequence-based, task-dependent models. We also examine reduced-channel configurations (16, 8, and 4 channels) to illustrate feasibility for routine clinical deployment and discuss how attention weights can provide insight into the relative contributions of different modalities and channels.

### 2.7.3 Statistical Analysis

For model evaluation, we conducted one-way ANOVAs to test whether performance differed across model modalities. In each ANOVA, model modality (EEG-only, descriptive-variable–only, and multi-modal) was the independent categorical factor, and a single performance metric (accuracy, recall, precision, F1-score, or Cohen's kappa) served as the dependent continuous variable.

Before conducting ANOVA, we assessed the normality of the performance distributions for each metric and modality using the Shapiro–Wilk test; no significant deviations from normality were observed (all $p > 0.05$), supporting the use of parametric ANOVA. ANOVA comparisons were conducted on the full-channel models (BiSAM-63) only, with model modality (EEG signal–only, descriptive-variables–only, and multi-modal) as the independent factor. Reduced-channel variants (BiSAM-16, BiSAM-8, BiSAM-4) were excluded from ANOVA because they are defined only for the EEG-based and multi-modal configurations and therefore do not constitute a complete level of the modality factor; these models were evaluated descriptively. All ANOVAs were performed on performance metrics obtained from the primary binary classification task (PDFOG+ vs combined PDFOG– + healthy controls). Secondary label configurations were analyzed exploratorily and reported descriptively.

All dependent variables in the statistical analyses were model performance metrics: accuracy, recall, precision, F1-score, and Cohen's kappa. Because these unitless measures are defined on comparable bounded scales, no additional normalization was applied for these dependent variables.

During significance testing, all performance metrics were computed on the fixed subject-wise test set across 90 training epochs for each modality configuration. These epoch-wise values served as repeated observations and were used to estimate mean performance for subsequent statistical analysis. ANOVAs were then conducted to assess whether performance differed significantly among the three modality configurations (EEG-only, descriptive-variable–only, and multimodal). When a significant main effect was observed ($p < 0.05$), Tukey's Honest Significant Difference (HSD) post-hoc test was applied to identify which modality pairs showed statistically significant differences in performance.

## 3. Results

Our primary analysis focused on a binary classification task contrasting PDFOG+ with the combined PDFOG– + healthy control group (PDFOG– + HC), and all results in this section refer to this main comparison unless otherwise specified. Table 2 compares the performances of primary BiSAM models using different input types. The EEG-only model with all 63 channels (BiSAM-63) achieved the best results in its class, with 76% accuracy and 76% recall. In contrast,

EEG-only models with fewer channels (BiSAM-16, BiSAM-8, BiSAM-4) showed much lower accuracy and recall (Table 2). The reduced-channel models appeared unstable since their limited EEG input left far less information for the classifier. The descriptive variables-only model (BiSAM-DV) achieved 68% accuracy and a balanced F1-score of 60%, indicating moderate discriminative power from descriptive variables features alone. By far the highest performance was seen in the multi-modal models that combine EEG and descriptive inputs. For example, the multi-modal BiSAM models using 8 or 4 EEG channels reached 88% accuracy each, with high F1-scores (88% and 83%, respectively). These values surpass all unimodal model scores and demonstrate that integrating both signal types yields substantially better classification.

The poor performance of the low-channel EEG-only models suggests that reducing EEG channels removes critical information. The full 63-channel model likely captures a broad range of neural activity, whereas trimming down to 16, 8, or 4 channels leaves the classifier with insufficient signal variance. In other words, the EEG-only models were "unstable" when given sparse input data. This instability likely accounts for their lower accuracy and recall relative to the full-channel model and the multimodal models. Similarly, the descriptive-only model showed only moderate effectiveness: its 68% accuracy and 60% F1-score indicate that objective descriptive features carry some relevant information but are not as rich or consistent as the EEG signals.

In contrast, the multi-modal models consistently outperformed both unimodal variants. Combining EEG with descriptive variables features appears to introduce greater variance and complementary information into the classifier. In our results, the BiSAM models that merged EEG and descriptive variables inputs showed markedly higher accuracy and F1 scores. This suggests that the EEG signals and objective descriptive features complement each other, providing a richer feature set. Consistent with prior work showing that multimodal fusion captures complex patterns more effectively than unimodal inputs (Kang *et al.*, 2023), this likely explains why the multimodal BiSAM variants achieved the strongest overall performance.

Moreover, we evaluated Cohen's kappa as a reliability metric for classification. Only the multi-modal models produced Kappa values above 0.5, whereas all unimodal (EEG-only or descriptive variables-only) models fell below this threshold. In standard interpretation, Kappa $> 0.5$ indicates at least moderate agreement. Thus, only the multi-modal BiSAM classifiers demonstrated consistently reliable predictions beyond chance. In line with previous research (McHugh, 2012; Dolatshahi *et al.*, 2021), achieving kappa above 0.5 supports the conclusion that the classifications of multi-modal models are meaningfully reliable. The corresponding Cohen's kappa values for all models are summarized in Table 2.

In the secondary PD-only analysis (PDFOG+/PDFOG–/Healthy Controls; see Table 3), the EEG-only BiSAM achieved an accuracy of 58.82% with kappa score 0.20, indicating fair agreement between predicted and true labels beyond chance, while the multimodal variant reached 71% with kappa score 0.50, indicating moderate-to-substantial agreement beyond chance. The descriptors-only model achieved an accuracy of 52.9% with kappa score 0.06, indicating only slight agreement between predicted and true labels beyond chance. Although the model correctly identified some patterns associated with FOG, its performance suggests that when trained using only descriptive features, performance remained close to chance.

For the three-class analysis (PDFOG+/PDFOG–/Healthy Controls), the multimodal model achieved 72% overall accuracy with kappa score 0.58 (substantial agreement beyond chance); the descriptors-only model achieved 60% accuracy with kappa score 0.39 (moderate agreement), and the EEG-only model achieved 40% accuracy with kappa score 0.01 (no reliable agreement).

For model evaluation, based on the statistical test, for the primary analysis, the one-way ANOVAs (Table 4) revealed highly significant differences across modalities for all performance metrics including accuracy ($F(2,87)=57.22$, $p<0.001$), recall ($F(2,87)=57.11$, $p<0.001$), precision ($F(2,87)=44.92$, $p<0.001$), F1-score ($F(2,87)=91.59$, $p<0.001$), and Cohen's kappa ($F(2,87)=120.67$, $p<0.001$). For secondary PD-only analysis, ANOVA showed highly significant differences among the three modalities: EEG-only, descriptive variable-only and multi-modal, across all evaluated metrics: accuracy ($F(2, 87)=490.05$, $p<0.001$), recall ($F(2, 87)=589.50$, $p<0.001$), precision ($F(2, 87)=543.14$, $p<0.001$), F1-score ($F(2, 87)=622.87$, $p<0.001$), and Cohen's kappa ($F(2, 87)=5242.03$, $p<0.001$). For secondary, 3-class setting analysis, ANOVA showed highly significant differences among the three modality configurations across all performance metrics: accuracy ($F(2,87)=2141.31$, $p<0.001$), recall ($F(2,87)=1876.43$, $p<0.001$), precision ($F(2,87)=1897.27$, $p<0.001$), F1-score ($F(2,87)=2479.12$, $p<0.001$), and Cohen's kappa ($F(2,87)=9308.81$, $p<0.001$). In all cases, post-hoc tests indicated that the multi-modal model performed significantly better than both the EEG-only and descriptive variable models across all metrics ($p<0.05$).

## 4. Discussion

In this study, we utilized the BiSAM architecture across three types of data configurations, signal-only, descriptive variables-only, and multi-modal, to assess their ability to classify PDFOG+. We do not propose resting-state EEG as a replacement for clinician-observed benchmarks (e.g., video-rated percent time frozen or MDS-UPDRS III item 11); rather, it offers an objective, neurophysiological complement, particularly valuable when gait provocation is unsafe or impractical, and may help triage patients, support longitudinal monitoring, or inform rehabilitation/neuromodulation planning. The input size and sampling framework remained consistent across all models and subgroups, allowing for an unbiased comparison of performance across varying data modalities and channel configurations. Our findings clearly demonstrate the limitations of the signal-only models when fewer EEG channels were used. As the number of channels decreased, classification accuracy and F1-scores dropped significantly. This reduction in performance is likely due to the imbalance of true positive cases in the dataset, which reduced the ability of models to generalize patterns effectively. Specifically, signal-only models struggled to learn robust discriminative features from the EEG signals when channel information was sparse, especially in identifying true positives. This is consistent with prior research showing that a reduced number of EEG channels can limit spatiotemporal resolution and degrade performance in motor classification tasks (Cassani *et al.*, 2018).

Conversely, the multi-modal models consistently outperformed both unimodal configurations across all channel settings. The integration of descriptive variables with EEG signals provided the models with both neurophysiological and contextual patient-specific information (Li *et al.*, 2016; Liu *et al.*, 2024), allowing for more precise detection of gait dysfunction. Even when the number of EEG channels was reduced to just eight, the multi-modal models (BiSAM-8 and BiSAM-4) demonstrated high classification performance, with BiSAM-8 achieving the best overall results across accuracy, precision, recall, and F1-score. The higher accuracy of the multimodal BiSAM-8 and BiSAM-4 models compared to the full EEG-only model is supported by their higher Cohen's kappa values, indicating stronger agreement between predicted and true class labels beyond chance. Removing less relevant channels reduces the dilute in the feature set that gives the models the ability to work on a more focused feature set. These results support findings from other research highlighting that multi-modal ML approaches outperform unimodal classifiers by capturing complementary data dimensions (Huang *et al.*, 2023; Jiang *et al.*, 2024).

In both the PD-only (PDFOG+ vs PDFOG–) and three-class settings, the pattern was consistent: EEG-only models were informative but limited (e.g., PD-only accuracy 58.8%, kappa score 0.20), descriptors-only models were near chance (52.9%, kappa score 0.06), and multimodal fusion yielded the strongest and most reliable agreement beyond chance (PD-only 71.0%, kappa score 0.50; three-class 72.0%, kappa score 0.58). These results indicate that EEG carries FOG-related signal, demographic/clinical variables add weak complementary cues, and combining both improves discrimination without relying on PD-vs-healthy differences.

Evaluation using ANOVA statistical analysis further validated these observations. According to the ANOVA test, these findings confirm that integrating EEG signals with descriptive features substantially enhances classification performance, yielding higher accuracy, balanced predictive capability (F1), and stronger agreement beyond chance (Kappa) compared to single-modality approaches.

Our comprehensive analysis suggests that the time and resource burden of EEG acquisition in clinical environments can be reduced without sacrificing model performance. The BiSAM-8 model, for example, retained exceptional classification ability even after EEG channels were reduced from 63 to just 8. The top contributing channels, TP9, FT8, Oz, Fp1, POz, C1, Iz, and T8, were consistently identified as the most informative for PDFOG+ detection. These results align with prior studies that emphasize the importance of temporoparietal and occipital regions in postural control and gait regulation in PD (Nwogo *et al.*, 2022; Jaramillo-Jimenez *et al.*, 2023).

Temporoparietal features are neurobiologically plausible in FOG because these regions support multisensory (visual–vestibular–proprioceptive) integration and spatial attention essential for balance and gait control. In line with prior EEG work on motor control, our model operates on power in lower-to-mid frequency bands (theta, alpha, beta) as well as gamma, with these bands selected based on their reported relevance to PD and FOG. Potential muscle contamination at TP electrodes was mitigated by the resting-state protocol and a rigorous preprocessing pipeline (band-pass and notch filtering, ICA-based removal of ocular/muscle components, artifact rejection, and re-referencing). Furthermore, the similar performance of reduced-channel configurations relative to the full montage indicates robustness and suggests that the classifier does not depend on any single TP electrode. A targeted exclusion analysis of specific TP channels is a promising direction for future work.

Traditional assessments like the MDS-UPDRS and patient interviews, while commonly used in PD evaluations, suffer from subjectivity and limited reproducibility. Clinical bias and ordinal scoring further reduce diagnostic reliability (Shulman *et al.*, 2010). In contrast, objective patient characteristics such as age, disease duration, and education level offer stable, quantitative descriptors that can meaningfully contribute to model interpretability and prediction accuracy. These features have been correlated with motor and cognitive outcomes in PD and may serve as essential indicators when designing individualized treatment strategies (Park *et al.*, 2025). The inclusion of EEG-derived features also plays a crucial role in capturing the neural basis of gait abnormalities. Previous research has identified altered beta-band synchronization and reduced connectivity in motor and frontal regions as biomarkers of FOG and other motor impairments in PD (Olde Dubbelink *et al.*, 2014; Singh *et al.*, 2020). EEG data thus provide high temporal resolution information that augments the spatial and semantic insights captured by clinical and demographic variables.

Prior work on this dataset reported LR AUC-ROC = 0.63 and LSTM AUC-ROC = 0.68 (accuracy 0.63) using Cz/Cz-cluster EEG and cross-validated protocols (Roy *et al.*, 2025). Our approach differs in two key respects: (i) it targets resting-state EEG with a bi-cephalic self-attention architecture that fuses EEG with objective descriptive variables, and (ii) it emphasizes clinical feasibility via reduced-channel configurations and a fixed subject-wise, group-stratified train/validation/test split to assess generalization to unseen participants. Because evaluation protocols (k-fold cross-validation vs. fixed subject-wise split) and channel sets can shift absolute performance values, we report our accuracy under this fixed split and interpret it relative to the above ranges rather than claiming direct superiority. This positions BiSAM as a complementary, clinic-oriented alternative to sequence-based models such as LSTM.

Our findings are directly relevant to the freezing population in three ways. First, clinical utility: a resting-state EEG–only classifier allows FOG assessment when gait tasks are infeasible (e.g., severe motor impairment, fall risk), offering an objective complement to questionnaires and IMU scores. This supports clinic triage (identifying PD patients likely to exhibit FOG), patient stratification for targeted rehabilitation or neuromodulation trials, and longitudinal monitoring of FOG burden without requiring task performance. The reduced-channel models improve clinical feasibility, enabling shorter recordings and simpler setups. Second, contribution to the literature: whereas most IMU studies emphasize task-based detection (and inconclusive prediction), our results demonstrate that resting-state cortical oscillations carry discriminative information for PDFOG+ vs. PDFOG–, and that multimodal fusion (EEG + descriptive variables) yields further gains. The selected band implicates specific frequency components, providing hypotheses about pathophysiology that can guide targeted interventions. Third, future translation: prospective, multi-site validation, calibration for decision thresholds, and integration with IMUs could enable hybrid toolkits that combine neurophysiology with movement sensing to support prevention and individualized therapy.

EEG abnormalities relevant to our features are well characterized on this cohort: PD vs. healthy controls shows reduced 1–4 Hz and increased 4–50 Hz power (Anjum *et al.*, 2020; Anjum *et al.*, 2024). Within PD, freezing of gait is most consistently associated with midfrontal beta activity, with additional contributions from theta/alpha in attention/postural networks (Roy *et al.*, 2025). Our classifier uses band-limited power in theta, alpha, beta, and gamma as inputs; these bands were selected based on prior evidence implicating them in PD and FOG (Singh, 2018; Singh *et al.*, 2020), but the present study does not estimate their relative discriminative contribution.

Despite the promising results, this study has several limitations. The use of resting-state EEG, while informative, may not fully capture task-specific neural dynamics associated with gait initiation or freezing episodes. Incorporating movement-related or task-based EEG recordings in future studies may yield more behaviorally relevant features (Borzi *et al.*, 2023). The identified set of eight optimal EEG channels was derived from this specific dataset and may not generalize across different populations or EEG systems. Therefore, external validation using diverse cohorts and standardized hardware is essential to confirm the reproducibility of the channel selection strategy. Although the BiSAM model achieved high performance, its computational complexity may limit real-time clinical applicability. Future work should explore more lightweight or deployable versions of the model to enhance its usability in clinical settings. While objective variables such as age, education and disease duration improved model accuracy, other potentially relevant factors, like medication status, or genetic markers, were not included. Integrating these could further improve model interpretability and predictive power (Makarious *et al.*, 2022; Huang *et al.*, 2023). A formal head-to-head benchmark against additional baselines (e.g., SVM, LDA) under a harmonized protocol is outside the present scope and will be pursued in follow-up work. Another study constraint is that recordings were obtained only in the ON state. As a result, the model characterizes ON-state FOG and may not generalize to levodopa-responsive (OFF-only) freezing. This introduces potential selection bias toward patients with ON-state FOG. Future work should include standardized OFF assessments or formal stratification by levodopa responsiveness and medication state to evaluate generalizability.

Our results align with the growing consensus that multi-modal data fusion, combining neurophysiological signals and clinical data, provides a richer and more reliable foundation for ML in neurodegenerative disease classification (Makarious *et al.*, 2022; Jiang *et al.*, 2024; Park *et al.*, 2025). Multi-modal fusion is known to improve performance in related tasks: for example, a recent emotion-recognition study found that adding EEG to video and audio modalities yielded more robust predictions than any single source alone (Liu *et al.*, 2024). Similarly, another study reported that an EEG+fNIRS hybrid system improved accuracy by about 12% over using EEG alone (Deligani *et al.*, 2021). In the current study, channel-level importance is used operationally to design reduced montages that preserve accuracy while improving clinical feasibility; it is not interpreted causally. For future evaluations, we will pre-specify a minimal montage based on these ranks and assess its stability in prospective cohorts.

By refining input features and minimizing channel usage, our approach offers promising directions for scalable, real-world deployment in both clinical and remote-monitoring settings. Future work will focus on validating the model with larger, longitudinal datasets and exploring its potential in real-time applications. Overall, this study contributes to the development of objective, data-driven tools that can enhance early detection and personalized monitoring of PD-related gait dysfunction.

## Author Contributions

Shomoita Jahid Mitin: conceptualization, data curation, formal analysis, investigation, methodology, project administration, software, visualization, writing – original draft.

Rodrigue Rizk: conceptualization, investigation, methodology, project administration, software, visualization, writing – review and editing.

Maximilian Scherer: conceptualization, methodology, writing – review and editing.

Thomas Koeglsperger: conceptualization, methodology, writing – review and editing.

Daneil Lench: conceptualization, methodology, writing – review and editing.

KC Santosh: conceptualization, investigation, methodology, project administration, software, visualization, writing – review and editing.

Arun Singh: conceptualization, funding acquisition, investigation, methodology, resources, software, supervision, writing – review and editing.

## Acknowledgements

We would like to thank the Seed for Success Foundation for their support of our research.

## Conflicts of Interest

The authors declare no conflicts of interest.

## Data Availability Statement

Our dataset will be made available upon reasonable request.

## Peer Review

The peer review history for this article is available at

**Table 1: Demographic information of all participants.**

| | Age (years) | Schooling (years) | Disease Duration (years) | MOCA | FOGQ | UPDRS III | LEDD (mg) |
|---|---|---|---|---|---|---|---|
| **Healthy Controls (n=41)** | 71.3 ± 7.6 | 16.6 ± 2.2 | NA | 26.6 ± 2 | NA | NA | NA |
| **PDFOG– (n=41)** | 68.2 ± 7.6 | 15.1 ± 3.5 | 4.2 ± 3.2 | 25.3 ± 3 | 1.56 ± 1.3 | 9.54 ± 5.5 | 722 ± 412 |
| **PDFOG+ (n=42)** | 69.0 ± 8.3 | 15.6 ± 3.2 | 5.9 ± 4.4 | 23.6 ± 4 | 11 ± 4.2 | 17.21 ± 6.5 | 1003 ± 459 |

Values represent the mean and standard deviation.
MOCA: Montreal Cognitive Assessment; FOGQ: Freezing of Gait Questionnaire Assessment; UPDRS III: motor examination section of the Unified Parkinson's Disease Rating Scale; LEDD: Levodopa Equivalent Daily Dose.

**Table 2: Comparison of all models based on accuracy, recall, precision, F1-score and Kappa scores for primary analysis.**

| Group Comparison | Modality | Model | Accuracy | Recall | Precision | F1-Score | Kappa Score |
|---|---|---|---|---|---|---|---|
| Healthy Controls + PDFOG– vs PDFOG+ | EEG Signals | BiSAM-63 | 76% | 76% | 58% | 65% | 0.48 |
| | | BiSAM-16 | 28% | 28% | 82% | 17% | -0.09 |
| | | BiSAM-8 | 48% | 48% | 55% | 51% | -0.03 |
| | | BiSAM-4 | 55% | 55% | 30% | 39% | 0.07 |
| Healthy Controls + PDFOG– vs PDFOG+ | Descrip-tive Variables | BiSAM-DV | 68% | 60% | 60% | 60% | 0.33 |
| Healthy Controls + PDFOG– vs PDFOG+ | Multi-Modal | BiSAM-63 | 80% | 80% | 81% | 80% | 0.60 |
| | | BiSAM-16 | 72% | 72% | 77% | 74% | 0.56 |
| | | BiSAM-8 | 88% | 88% | 88% | 88% | 0.76 |
| | | BiSAM-4 | 88% | 81% | 85% | 83% | 0.74 |

**Table 3: Comparison of all models based on accuracy, recall, precision, F1-score and Kappa scores for secondary analysis.**

| **Group Comparison** | **Modality** | **Model** | **Accuracy** | **Recall** | **Precision** | **F1-score** | **Kappa score** |
|---|---|---|---|---|---|---|---|
| **PDFOG– vs PDFOG+** | **EEG Signals** | | 58.82% | 61% | 64% | 57% | 0.20 |
| | **Descriptive Variables** | BiSAM-DV | 52.9% | 54% | 54% | 54% | 0.06 |
| | **Multi-Modal** | | 71% | 72% | 73% | 71% | 0.57 |
| **Healthy Controls vs PDFOG– vs PDFOG+** | **EEG Signals** | | 40% | 40% | 41% | 40% | 0.01 |
| | **Descriptive variables** | BiSAM-DV | 60% | 60% | 61% | 61% | 0.40 |
| | **Multi-Modal** | | 72% | 73% | 73% | 73% | 0.58 |

**Table 4: One-way ANOVA results assessing the effect of model modality on classification performance metrics across binary and multi-class classification tasks for full models.**

| Group Comparison | Independent Variables: Modality | Dependent Variable: Metric | F values | p values |
|---|---|---|---|---|
| **PDFOG–<br>+<br>Healthy Controls<br>vs<br>PDFOG+** | **EEG-Only<br>vs<br>Descriptive Variable-Only<br>vs<br>Multi-modal** | Accuracy | 57.22 | <0.0001 |
| | | Recall | 57.11 | <0.0001 |
| | | Precision | 44.92 | <0.0001 |
| | | F1 Score | 91.59 | <0.0001 |
| | | Kappa Score | 120.67 | <0.0001 |
| **PDFOG–<br>vs<br>PDFOG+** | **EEG-Only<br>vs<br>Descriptive Variable-Only<br>vs<br>Multi-modal** | Accuracy | 490.05 | <0.0001 |
| | | Recall | 589.50 | <0.0001 |
| | | Precision | 543.14 | <0.0001 |
| | | F1 Score | 622.87 | <0.0001 |
| | | Kappa Score | 5242.03 | <0.0001 |
| **Healthy Controls<br>vs<br>PDFOG–<br>vs<br>PDFOG+** | **EEG-Only<br>vs<br>Descriptive Variable-Only<br>vs<br>Multi-modal** | Accuracy | 2141.31 | <0.0001 |
| | | Recall | 1876.42 | <0.0001 |
| | | Precision | 1897.27 | <0.0001 |
| | | F1 Score | 2479.12 | <0.0001 |
| | | Kappa Score | 9308.81 | <0.0001 |

## Figure Legends

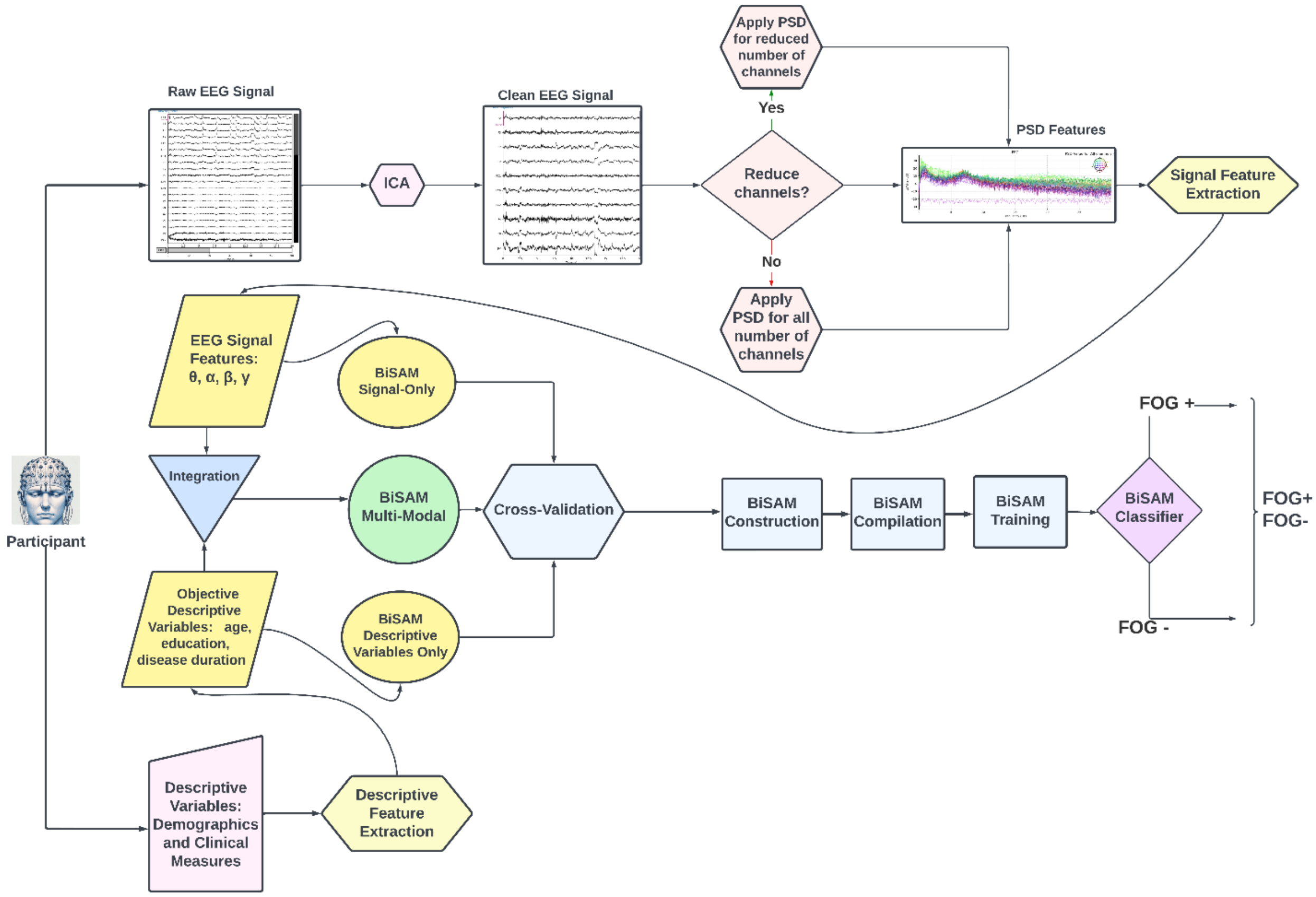

**Figure 1. Overview of the BiSAM modeling pipeline for primary analysis.** The diagram shows the end-to-end workflow for developing and evaluating the primary BiSAM model. Data were collected from participants, including EEG signals and descriptive variables (e.g., age, schooling, and disease duration). Three modeling configurations were implemented: signal-only, descriptive variables-only, and multi-modal (combining both feature sets). Each configuration was tested across different channel settings, including the full set of 63 EEG channels and reduced subsets (16, 8, and 4 channels), to assess the effect of dimensionality reduction on model performance.

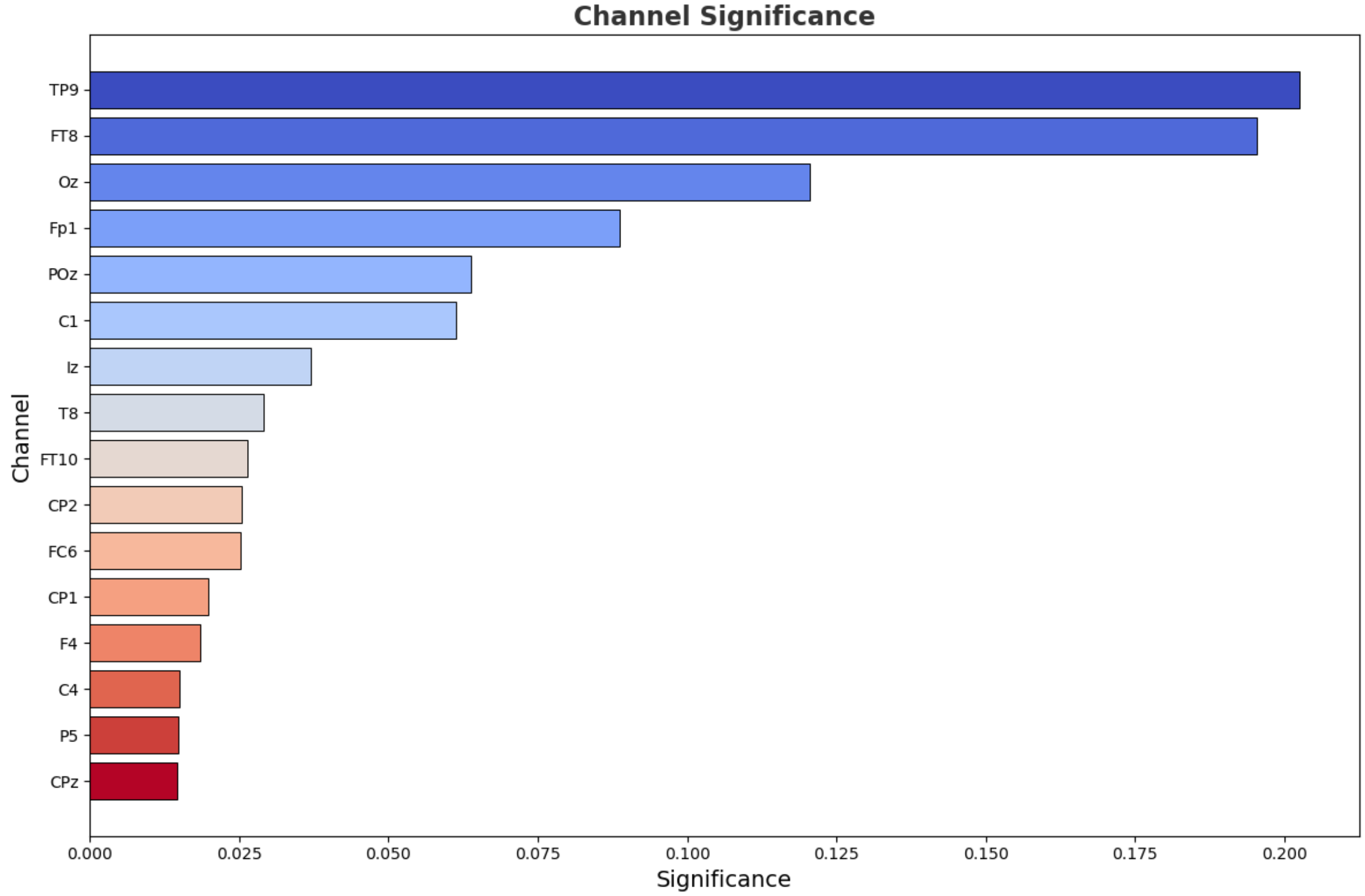

**Figure 2. Top-ranked EEG channels based on resting-state signals.** This figure displays the most significant EEG channels identified through feature importance analysis. The x-axis represents the importance scores assigned to each channel, while the y-axis lists the corresponding EEG channels. Channels are sorted in descending order based on their contribution to model performance, highlighting those most relevant for distinguishing between subject groups in the resting-state condition.

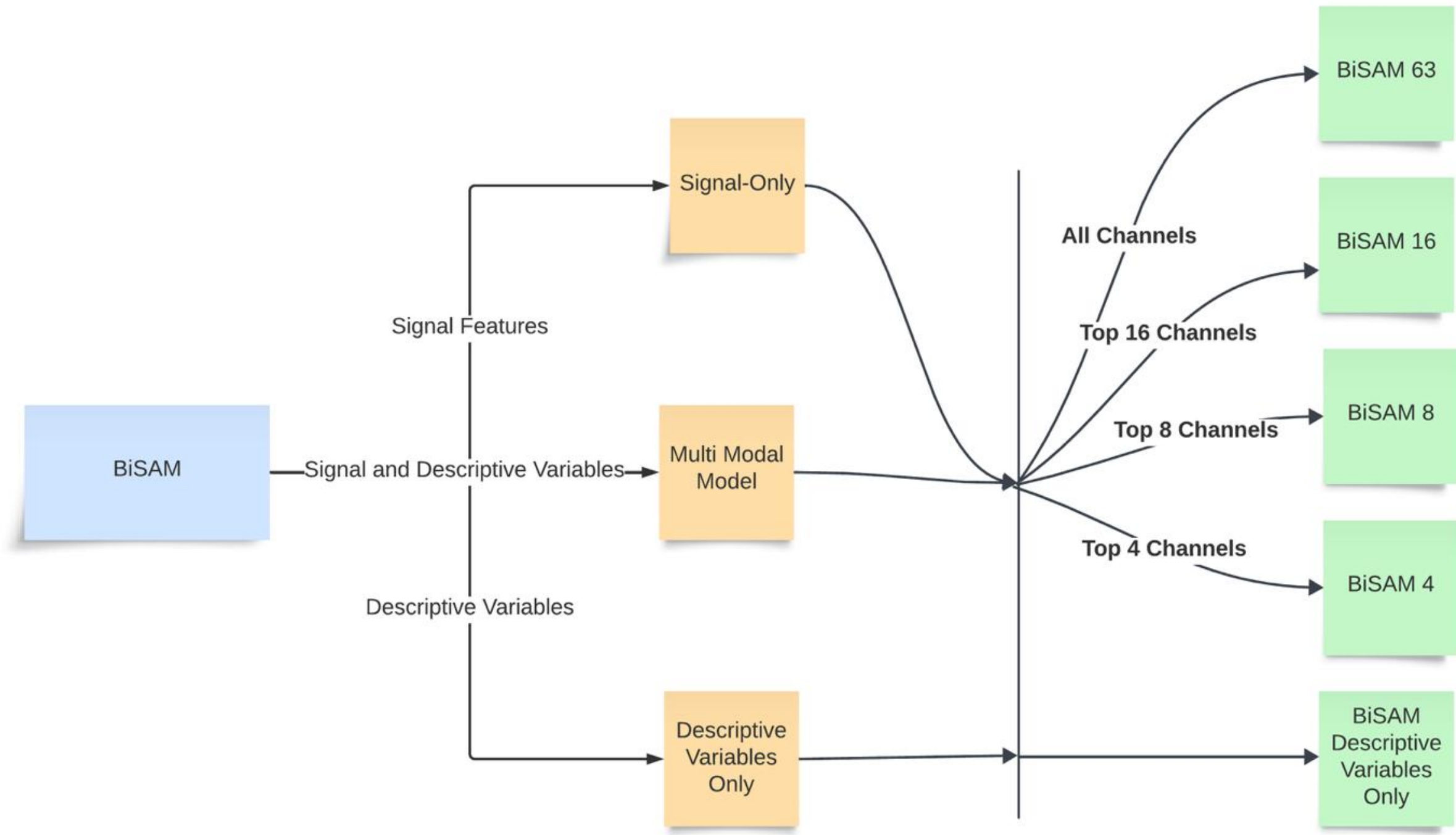

**Figure 3. Different categories of BiSAM model.** The BiSAM mechanism was applied to three different groups based on the selected features, including the signal-only modal, descriptive variables-only modal with only descriptive features and multi-modal for both signal and descriptive variable features, and later was categorized into four different subgroups based on the significance of the channels, including BiSAM-63, BiSAM-16, BiSAM-8, and BiSAM-4. Meanwhile, the significance of channels is purely related with signal data, so selected numbers of channels and their significance have no influence on the descriptive variables-only modal.

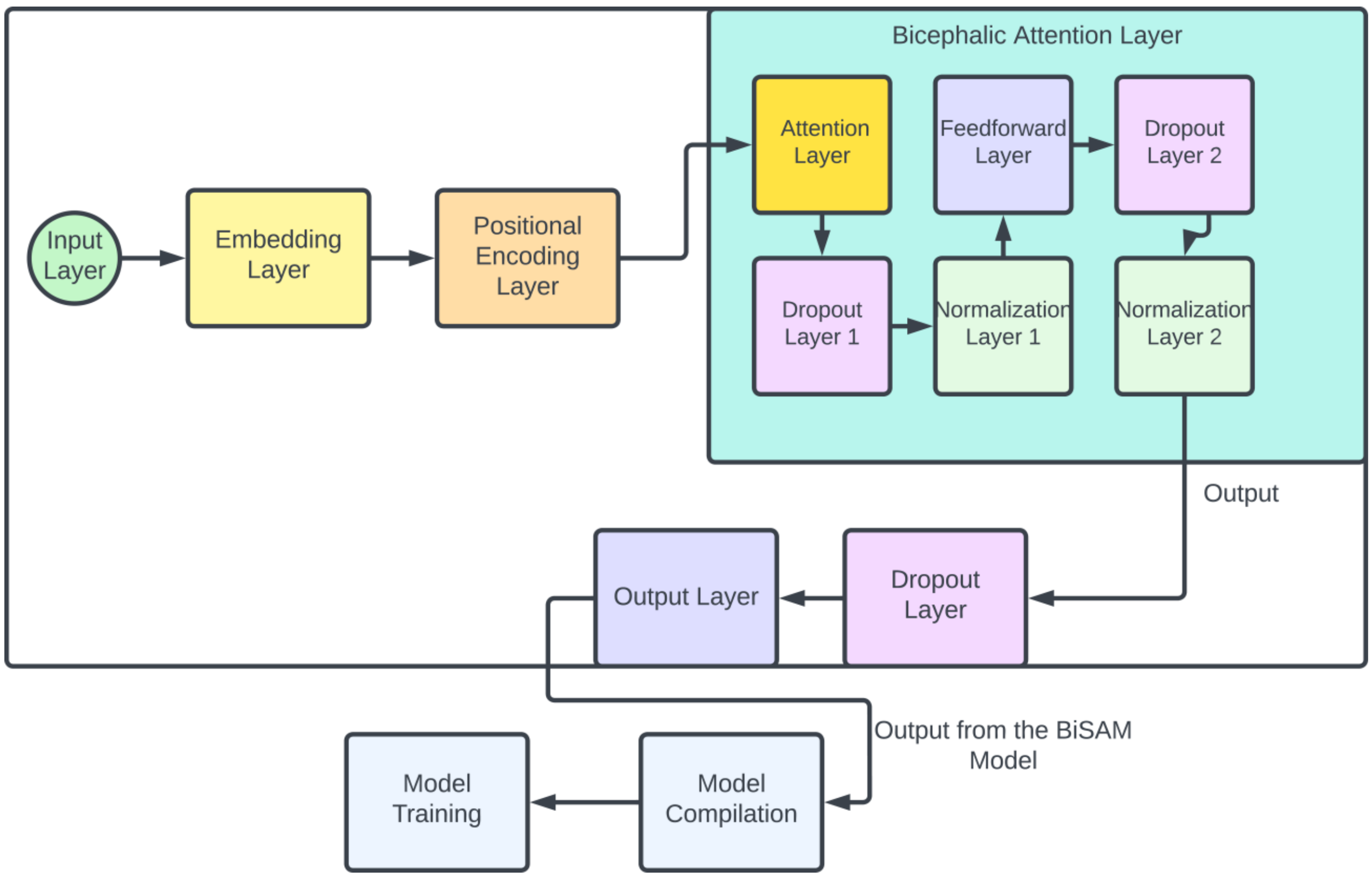

**Figure 4. BiSAM model architecture.** The architecture begins with an input layer representing either signal features, descriptive variables, or both, depending on the modality. The core of the model features a bi-cephalic structure, comprising two parallel attention pathways designed to process temporal and contextual patterns simultaneously. The final output layer performs the classification task, predicting subject group membership based on the learned representations.